\documentclass[journal,transmag]{IEEEtran}
\ifCLASSINFOpdf
\usepackage[pdftex]{graphicx}
\else
\fi

\usepackage{amssymb,amsmath}
\usepackage{latexsym}
\usepackage{url}
\usepackage{xcolor}
\definecolor{newcolor}{rgb}{.8,.349,.1}
\usepackage{algorithm}
\usepackage[noend]{algpseudocode}
\usepackage{bbm}
\usepackage{dsfont}

\def\eg{\emph{e.g.}}
\def\etal{et \emph{al.}}
\renewcommand{\Re}[1]{\mbox{$\mathbbm{R}^{#1}$}}

\newcommand{\tupleset}[1]{\mbox{${#1}^2$}}
\newcommand{\tuple}[1]{\mbox{$\mathbbm{#1}$}}
\newcommand{\sburst}[1]{\mbox{$\Phi_{#1}$}}
\def\indi#1{\mbox{$\mathds{1}_{#1}$}}

\begin{document}
	\title{Burst ranking for blind multi-image deblurring}
	
	%,~\IEEEmembership{Student,~IEEE}
	\author{\IEEEauthorblockN{Fidel A. Guerrero Pe\~{n}a\IEEEauthorrefmark{1},
			Pedro D. Marrero Fern\'{a}ndez\IEEEauthorrefmark{1},
			Tsang Ing Ren\IEEEauthorrefmark{1}, \\
			Jorge J. G. Leandro\IEEEauthorrefmark{2}, and
			Ricardo Nishihara\IEEEauthorrefmark{2}}
		
		\IEEEauthorblockA{\IEEEauthorrefmark{1}Centro de Inform\'{a}tica, Universidade Federal de Pernambuco, Brazil}
		\IEEEauthorblockA{\IEEEauthorrefmark{2}Motorola Mobility LLC (a Lenovo Company)}% <-this % stops an unwanted space
		\thanks{
			Corresponding author: Fidel Guerrero Pe\~{n}a (email: fagp@cin.ufpe.br).}}
	
	\markboth{Submitted to IEEE Transactions on Image Processing}%
	{Shell \MakeLowercase{\textit{et al.}}: Bare Demo of IEEEtran.cls for IEEE Transactions on Image Processing}

	\IEEEtitleabstractindextext{%
		\begin{abstract}
			We propose a new incremental aggregation algorithm for multi-image deblurring with automatic image selection. The primary motivation is that current bursts deblurring methods do not handle well situations in which misalignment or out-of-context frames are present in the burst. These real-life situations result in poor reconstructions or manual selection of the images that will be used to deblur. Automatically selecting best frames within the burst to improve the base reconstruction is challenging because the amount of possible images fusions is equal to the power set cardinal. Here, we approach the multi-image deblurring problem as a two steps process. First, we successfully learn a comparison function to rank a burst of images using a deep convolutional neural network. Then, an incremental Fourier burst accumulation with a reconstruction degradation mechanism is applied fusing only less blurred images that are sufficient to maximize the reconstruction quality. Experiments with the proposed algorithm have shown superior results when compared to other similar approaches, outperforming other methods described in the literature in previously described situations. We validate our findings on several synthetic and real datasets.
		\end{abstract}
		
		\begin{IEEEkeywords}
			Blurred Images Sorting, Deep Learning, Multi-image deblurring.
		\end{IEEEkeywords}}
		
		\maketitle
		
		\IEEEdisplaynontitleabstractindextext

		\IEEEpeerreviewmaketitle
		
		\section{Introduction}

\begin{figure*}[ht!]
    \begin{center}
        \centering
        \setlength{\tabcolsep}{1pt}
        \begin{tabular}{ccc}
            Original Burst&&Fourier Burst Accumulation\\
            \includegraphics[width=0.47\linewidth]{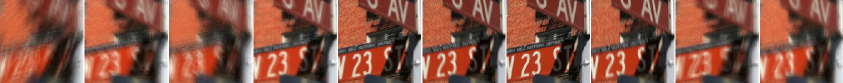}&$\rightarrow$&
            \includegraphics[width=0.47\linewidth]{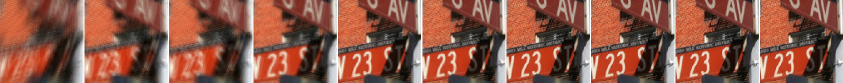}\\
            Sorted Burst (Ours)&&Incremental Fourier Burst Accumulation (Ours)\\
            \includegraphics[width=0.47\linewidth]{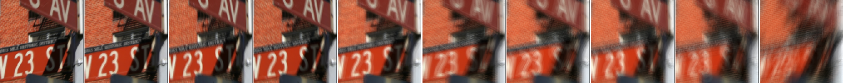}&$\rightarrow$&
            \includegraphics[width=0.47\linewidth]{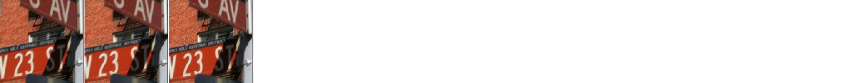}\\
        \end{tabular}
    \end{center}
    \caption{Original and sorted bursts with its respective aggregations. In our Incremental Fourier Burst Accumulation algorithm a degradation recognition mechanism interrupts the aggregation whenever a certain level of deterioration is observed during the reconstruction.}
    \label{fig:initial}
    %\vspace{-1cm}
\end{figure*}

Image captured by mobiles camera devices usually contain motion blur,  caused by hand tremors and dynamic scene content \cite{wieschollek2017learning}. The motion blur occurs because photons accumulation is the main principle of the image acquisition process. While more photons reach the sensor, a better image is obtained as the noise is minimized. However, a static scene is required because motion between the acquisition surface and the scene produce a wrong accumulation of photons on pixels neighborhood leading to a loss of sharpness.

Recent mobiles cameras capture a burst, which are a sequence of frames taken in a short period of time. Several researchers study the benefits of burst frames aggregations to form images with less noise \cite{buades2009note}.  However, the methods are affected by the blur in the frames. The multi-image deconvolution approach solves the inverse problem searching for the kernels and the latent sharp image \cite{zhang2013multi}. Although good results are obtained with this approach, it is very slow and have a high dependency on the kernel estimation method. This is a very a challenging problem by itself.

Because motion in mobile devices is originated from hand tremor, the nature of the blur is random \cite{carignan2010quantifying}. This implies that changes in a frame of the burst are independent to changes in others frames. Then, the motion blur in every image within the burst is also different. An information aggregation approach can be applied by taking a less blurred region of each image to build a sharper reconstruction \cite{delbracio2015removing}. The problem is still challenging as the less blurred region needs to be identified in each frame and, even after the correct identification, images might be misaligned.

Most of the modern methods for multi-frame deblurring require the input to match a fixed temporal size \cite{chakrabarti2016neural,wieschollek2016end,noroozi2017motion}. In these methods, it is established the burst length, and the aggregation process does not handle more extended frames sequences. This problem relaxation can significantly harm the deblurring task under real-life conditions because motion blur random nature does not guarantee that enough information is available to obtain a proper reconstruction given a fixed amount of frames.

To work with variable burst lengths, some authors performed a burst-size independent aggregation using a weighted average in the Fourier domain \cite{delbracio2015removing} and a recurrent network with temporal feature transference \cite{wieschollek2017learning}. Despite the advantages of these approaches, referred works do not consider that some images in the burst present severe blur degradation, misalignment (even after registration) or out-of-context frames that can lead to noisy reconstruction. We show experimentally that sorting the burst from less to higher blurred images and then applying an incremental aggregation with automatic frame selection will result in better image deblurring.

Our approach performs a local relative ranking of frames through a novel blur estimation bi-variable function approximated by a VGG \cite{simonyan2014very}, a well-known CNN model architecture. In order to train this model, we successfully generate a synthetic dataset of blurred images through motion blur kernels synthetically created with no cost in the acquisition process. In the sequel, the dataset samples annotation is also automatically carried out through the proposed kernel complexity metric score. Lastly, an incremental aggregation method is applied by using only two images at each step until the maximum quality reconstruction is obtained (equivalently a reconstruction degradation is measured) or no more images are left in the burst. The proposed algorithm deals with arbitrary lengths of sequences and performs an automatic relevant images selection. Fig. \ref{fig:initial} shows an example of a sorted burst and incremental aggregation.

The remaining of the paper is organized as follows. Section \ref{sec:related} discusses the related work and the substantial differences to what we propose. Section \ref{sec:method} explains how a burst can be sorted and fused while in Section \ref{sec:results} we present and discuss results of the proposed ranking function and aggregation procedure, comparing the algorithm to other literature methods. Conclusions are finally summarized in Section \ref{sec:conclusion}.

		\section{Related Work}
\label{sec:related}
\textbf{Image burst ranking:}
An approach closely related to our blur comparison function was proposed in \cite{zagoruyko2015learning} to learn a similarity function. Their study was conducted with several networks to regress a similarity score. The referred work shows that even without a direct notion of a descriptor, a simple VGG with two inputs corresponding to 8 bits images is sufficient to learn a comparison function. Here we extend the approach for classification over RGB images, and we force the network to meet the trichotomy law for better results.

Another method close to this was proposed in \cite{liu2013no} for automatic selection of optimal deblurring algorithms. The authors performed a massive user study and manually select the features that capture common deblurring artifacts. Then, a metric is learned combining the information of each computed feature to meet the rank order given by the users. Finally, a robust evaluation metric was introduced to compare the ranks obtained by their method and the ground truth. Their algorithm returned a global score but involve a blurry reference image in the metric computation. Although the problem solved here does not use any reference image, several ideas were taken from their work for sorting and evaluation. In another hand, the relevant features herein are learned by the network and not hand-crafted.

\textbf{Blind multi-image deblurring:}
One of the most relevant work in multi-image deblurring is the Fourier Burst Accumulation \cite{delbracio2015removing} (FBA). The authors present an algorithm that aggregates a burst of images in the frequency domain, taking what is less blurred of each frame to build an image that is sharper and less noisy than all burst frames. To this purpose, the method takes as input a series of registered images and computes a weighted average of the Fourier coefficients of the images. The algorithm is straightforward to implement and conceptually simple with excellent results. Despite the method usefulness, frames are assumed to be well registered (even those with to much blur) something very hard to obtain. The method proposed in here deals with this problem introducing a new degradation recognition mechanism. 

Wieschollek \etal \cite{wieschollek2017learning} recently proposed a novel U-Net based Recurrent Deblur Network (RDN). The network can handle arbitrary temporal sizes and propagate previous frames information through a new kind of temporal skip connections. Also, the information aggregation is performed incrementally like in this work. However, images where the scene changes drastically by motion blur or out of context are also considered during the deblurring reducing the reconstruction quality. Our proposed degradation recognition mechanism is used to handle this situation. Because their work uses an incremental aggregation network, we plan to extend our method using an RDN as the deblurring step.

		\section{Proposed Method}
\label{sec:method}

As mentioned above, we formulate the multi-image deblurring problem as a two steps process where it is performed a burst sorting as a local relative ranking task, and then an incremental aggregation method is applied. 

For the burst reorganization step, given a burst with $l$ frames, denoted as $S=\{x_0, ..., x_l\}$, the set of all possible pairs of $S$ is obtained as the cartesian product $S \times S$, also written as $S^2$. Let $\tuple{x}=(x_i,x_j)$ be a generic tuple of \tupleset{S}. Our goal is to find a comparison function $f_c(\tuple{x})$ that takes two frames of $S$ as input, and outputs the probability that $x_i$ is blurrier than $x_j$, $f_c \colon \tupleset{S} \rightarrow [0,1] \subset \Re{} $. From $f_c$ it can be defined the {\it blurrest} binary relation $R_{S}$ over $S$ using the maximum a {\it posteriori} decision rule for a binary classification problem, that is, $R_{S} = \{ \tuple{x} \left|\right. \text{P}(\tuple{x}) \geq 0.5, \forall \tuple{x} \in \tupleset{S}\}$, once $\text{P}(\tuple{x})$ has been directly computed through the $f_c$ output value. Then, a totally ordered set $\sburst{S}$ can be defined over $S$ given the binary relation $R_{S}$, $\sburst{S}=(S,R_{S})$.

We propose to approach the information aggregation step in an iterative way. The objective is to find an optimal ordered subset, $\bar{\Phi}_{S}^{*} \subset \sburst{S}$, such that $\hat{x}$ is as close as possible to the ideal sharp image $x$, $\hat{x}$ being the reconstructed image from the frames $x_i^* \in \bar{\Phi}_{S}^{*}$. An incremental reconstruction over an ordered burst $\sburst{S}$ is defined here as $\hat{x}_{t} = g_{\omega,p}(x_t,\hat{x}_{t-1})$, where $0 \leq t\leq l$ is an iteration number, $x_t$ is the frame $t$ of the sorted burst $\sburst{S}$, and $\hat{x}_t$ is the obtained reconstruction at iteration $t$, with $\hat{x}_0=x_0$. Let $G$ be the set of all possible incremental reconstructions over $\sburst{S}$, $G=\{\hat{x}_t \left|\right. 0 \leq t\leq l\}$. An incremental aggregation function $g_{\omega,p}$ is defined in a manner that the deblurring is performed iteratively using a previously deblurred image.
%, $g_{\omega,p} \colon \sburst{S} \times G \rightarrow G$
Let $R_G$ be the {\it blurrest} binary relation over $G$, and $\sburst{G}$ the obtained totally ordered set, $\sburst{G}=(G,R_{G})$. The maximum of $\sburst{G}$ is then defined as the sharpest reconstructed image in $G$, $\hat{x}_U = \hat{x}_t$ such that $\hat{x}_t \in G$ and $(\hat{x}_i, \hat{x}_t) \in R_G$, $\forall \hat{x}_i \in G$. Then, the searched optimal ordered subset $\bar{\Phi}_{S}^{*}$ will be the one that contains the minimum amount of elements that are sufficient to generate the maximum of $\sburst{G}$.

\begin{figure}[!tb]
    \includegraphics[width=0.9\linewidth]{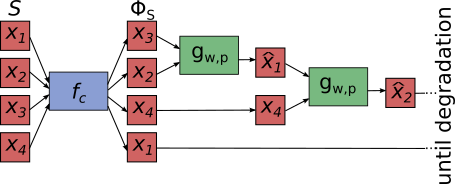}
    \caption{Example of the overall method scheme for an input burst $S$ of four frames sorted, and then performed an incremental aggregation deblurring with degradation recognition.}
    \label{fig:process}
\end{figure}

Fig. \ref{fig:process} depicts the overall process where burst sorting is the first step followed by an incremental aggregation with degradation recognition for deblurring. We perform the deblurring until a restoration improvement is no longer obtained. A detailed explanation is given below for every stage.

\subsection{Data set}

As will be discussed later, the target function $f_c$ is approximated through a CNN, and training such a neural network to predict the relative blur ranking given two blurry inputs requires realistic training data. Because there is no public burst dataset available for our ranking purpose, we synthetically generate our dataset to train the comparison function $f_c$. One way to do it is sampling continuous frames from existing videos to simulate burst sequences which seem to be reasonably straightforward. This approach was used in recent works \cite{su2017deep,noroozi2017motion} building a training dataset by recording videos captured at 240fps with a GoPro Hero camera. Next, the average is taken over obtained images providing an acceptable synthetic motion blur. However, a huge amount of blur needs to be manually assigned to each frame in a given tuple $\tuple{x}$ for the creation of the $R_S$ binary relation. We have found very expensive to create by hand the enormous amount of ground truth data required for the network. Another way to perform this task as accurate as possible requires also a sharp image of the scene to compute a blur score using a reference metric. Nevertheless, while a significant effort can be made to construct the dataset, there is a limitation in the number of captured videos, the used recording devices and in the diversity of scenes.

Based on the data generation proposed in \cite{wieschollek2016end} we generated our dataset by applying synthetic blur kernels to patches extracted from the MS COCO dataset \cite{lin2014microsoft}. This dataset is composed of highly variated real-world images obtained from the internet. For a fair evaluation, we use the provided splitting in training and validation set. Optimizing $f_c$ network parameters is done on the training set only. This kind of data generation gives us a nearly infinite quantity of training data.

Blur kernels were created following \cite{boracchi2012modeling} and using authors implementation. First, a particle random motion trajectory with length $m$ is generated. Then, the Point-Spread Function (PSF) $h$ is obtained by sampling the continuous trajectory on a regular pixel grid using subpixel linear interpolation, $h \colon \Omega \rightarrow \Re{}$ with $\Omega \subset \Re{2}$. A blurred input tuple $\tuple{x}=(x_0,x_1)$ is generated on-the-fly by applying two created kernels $h_0$ and $h_1$ to a randomly selected sharp patch $x$.

\subsection{Measuring blur}

\begin{figure*}[!htb]
    \begin{center}
        \centering
        \setlength{\tabcolsep}{1pt}
        \footnotesize
        \begin{tabular}{lllll}
            \includegraphics[width=0.19\linewidth]{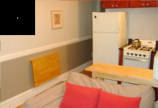}&
            \includegraphics[width=0.19\linewidth]{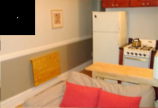}&
            \includegraphics[width=0.19\linewidth]{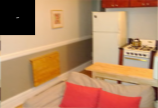}&
            \includegraphics[width=0.19\linewidth]{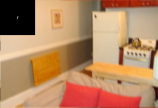}&
            \includegraphics[width=0.19\linewidth]{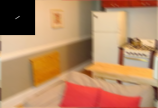}\\
            $C_l:0.62$, $C_s:0.50$, $m:3$ & $C_l:0.96$, $C_s:0.41$, $m:3$ & $C_l:1.49$, $C_s:0.39$, $m:5$ & $C_l:1.51$, $C_s:0.40$, $m:5$ & $C_l:2.65$, $C_s:0.42$, $m:9$\\
            $hm:0.27, MSE:0.0079$ & $hm:0.29, MSE:0.0080$ & $hm:0.31, mse:0.0080$& $hm:0.32, MSE:0.0086$& $hm:0.36, MSE:0.0094$\\
            $PSNR:21.01, SSIM:0.97$& $PSNR:20.98, SSIM:0.96$& $PSNR:20.97, SSIM:0.96$& $PSNR:20.68, SSIM:0.95$& $PSNR:20.28, SSIM:0.94$\\
            $\zeta \text{ (Ours)}: 0.01$&$\zeta \text{ (Ours)}: 0.02$&$\zeta \text{ (Ours)}: 0.03$&$\zeta \text{ (Ours)}: 0.04$&$\zeta \text{ (Ours)}: 0.16$\\
            
            \includegraphics[width=0.19\linewidth]{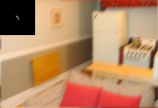}&
            \includegraphics[width=0.19\linewidth]{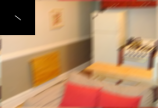}&
            \includegraphics[width=0.19\linewidth]{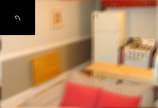}&
            \includegraphics[width=0.19\linewidth]{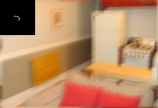}&
            \includegraphics[width=0.19\linewidth]{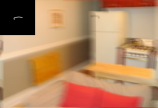}\\
            $C_l:2.91$, $C_s:0.61$, $m:19$ & $C_l:3.81$, $C_s:0.41$, $m:13$ & $C_l:3.32$, $C_s:1.77$, $m:19$ & $C_l:3.94$, $C_s:1.44$, $m:17$ & $C_l:5.01$, $C_s:1.04$, $m:19$\\
            $hm:0.50, MSE:0.0094$ & $hm:0.37, MSE:0.0100$ & $hm:1.16, MSE:0.0114$& $hm:1.06, MSE:0.0117$& $hm:0.86, MSE:0.0132$\\
            $PSNR:20.29, SSIM:0.93$& $PSNR:20.00, SSIM:0.93$& $PSNR:19.45, SSIM:0.91$& $PSNR:19.32, SSIM:0.91$& $PSNR:18.80, SSIM:0.91$\\
            $\zeta \text{ (Ours)}: 0.26$&$\zeta \text{ (Ours)}: 0.33$&$\zeta \text{ (Ours)}: 0.49$&$\zeta \text{ (Ours)}: 0.53$&$\zeta \text{ (Ours)}: 0.63$\\
        \end{tabular}
    \end{center}
    \caption{Generated burst manually ordered in an increasing blur order and different quality metrics for each image. Image blur increases from left to right and from top to bottom.}
    \label{fig:kernels}
    %\vspace{-1cm}
\end{figure*}

To measure the quantity of blur it is typical to use image quality metrics as shown in the Fig. \ref{fig:kernels}, where the notation $MSE$, $PSNR$ and $SSIM$ refers to the Mean Square Error, Peak Signal-to-Noise Ratio, and Structural Similarity Index quality metrics respectively. Despite these metrics detect the amount of blur as expected in cases as seen in the Fig. \ref{fig:kernels}, they are also affected by other deteriorations like noise, illuminance change and simple image shift. 

Because the amount of blur in an image depends solely on the trajectory, and we synthetically generate the kernels, a better blur metric can be obtained from it rather than computing a reference quality metric with the sharp patch $x$. Given two kernels $h_0$ and $h_1$, the goal is to compute a blur score over each kernel and compare them to create the binary relation ground truth, $R_S$. To this end, we explore several PSF descriptors as can be seen in Fig. \ref{fig:kernels}. For each image in the figure the respective values for the following quantities are presented: $m$, the trajectory length, $C_l$ and $C_s$ that corresponds to trajectory eigenvalues \cite{boracchi2012modeling}, and $hm$ that represent half of the harmonic mean between eigenvalues. From a quick inspection of the figure it can be observed that from the aforementioned metrics a burst ranking cannot be obtained as in the manually sorted ground truth. The disagreement occurs because the methods only focus either on the trajectory shape or the length for describing the movement complexity, without considering both merged in a single descriptor, or the shifting kernels effect over the metric. A shifted kernel refers to a PSF with non-vanishing first moment. That is, $\int_{k \in \Omega} h(k)\textit{d}k \neq 0$ \cite{delbracio2015removing}.

We observed that the farther the position in the trajectory from the kernel center, the worst is the blur because a broader pixel neighborhood is used in the degradation process. On the other hand, the larger the region, the higher the probability of averaging out of context pixels information. We propose a metric that intrinsically considers the trajectory length, shape and shift as an exponential distance between the kernel and its center.

\begin{equation}
\zeta_h= 100 \cdot \int_{k_1} \int_{k_2} h(k_1,k_2)  \left[1-e^{ - \frac{k_1^2 + k_2^2}{2\sigma^2} }\right] \textit{d}k_1\textit{d}k_2
\label{eq:kernel_complex}
\end{equation}
where $k=(k_1,k_2), k \in \Omega \subset \mathbb{R}^2$ and $\sigma$ is the greatest value of $m$ that can be obtained, in this work $\sigma=32$. In Fig. \ref{fig:kernels} it can be seen the proposed metric value and its correspondence with the ground truth order. Also, it is observed that similarly blurred images have closer values of $\zeta$ whereas high differences are perceptually differentiated. 

After computing $\zeta_{h_0}$ and $\zeta_{h_1}$ for a generated kernel pair $(h_0,h_1)$, a training tuple $\tuple{x}=(x_0,x_1)$ will belong to the {\it blurrest} binary relation, $\tuple{x} \in R_S $, {\bf iff} $\zeta_{h_0} > \zeta_{h_1}$, \eg\ the image $x_0$ is blurrier than $x_1$.

\subsection{Burst images ranking}

Although the proposed blur metric (Eq. \ref{eq:kernel_complex}) allows sorting a burst according to the blur of each image, it is restricted to cases when the PSF that originated the degradation is known. Then, as discussed in Sec. \ref{sec:method}, a function $f_c$ that can predict the amount of blur of an image within an unknown-length burst needs to be learned. In this work it is proposed the use of a convolutional neural network to determine both the discriminative features as well as the comparison function $f_c$ to be used in the ranking. Although the blur score $\zeta_h$ is known and can be used to train a regression function \cite{li2017evaluation}, it is very slow to train and context dependent. We also find that this type of Image Quality Assignment (IQA) models requires specific regression operations like pooling by weighted average patch aggregation \cite{bosse2018deep} and general regression neural networks \cite{yu2016cnn} that needs to be tuned depending on the metric to be learned. 

To circumvent these issues we solved the problem as a relative ranking measure. Then, the problem is cast as a binary classification task using a simple VGG architecture to approximate the trainable $f_c$ function. This method allows using any classification architecture for sorting a set of images. The network input is a pair of blurred images $\tuple{x}=(x_0,x_1)$ and the prediction is the amount of blur in each image relative to the other, \eg\ the probability that image $x_0$ is blurrier than $x_1$ and vice versa. For a given pair $\tuple{x}$ from the training set, the classification label $y$ used to train VGG is the indicator function $\indi{R_S}(\tuple{x})$ over the ground truth set $R_S$, \eg\ $y=1$ if $\tuple{x} \in R_S$, otherwise $y=0$. Using an input pair $\tuple{x}$ rather than directly regressing the blur score of one image simplifies the comparison task and remove images dependence as the network sees at once two frames that belong to the same burst and therefore, the same scene. In so doing, the features that make one image to looks blurrier than the other without considering the scene context.

Some authors use a VGG-based comparison network to perform images triage with Siameses \cite{chang2016automatic} and Generative Adversarial Networks \cite{wang2018real} architectures. Our blur comparison network, however, is closest to the approach studied in \cite{zagoruyko2015learning} to learn a similarity function because its superiority was validated concerning one input image-based approaches. Nevertheless, we use color images for a classification task rather than a grayscale image for regression. Here it is simply considered the two RGB images of an input pair $\tuple{x}$ as a 6-channel map, which is directly fed to the first convolutional layer of the VGG. As in \cite{zagoruyko2015learning}, there is no direct notion of a descriptor in the architecture. This network provides greater flexibility compared to the above models as it starts by processing the two images jointly. During network training the output vector $\tuple{z}=(z_0,z_1)$ is a $1 \times 2$ map used for Binary Cross Entropy (BCE) loss function minimization, such that  $\tuple{z} \approx (y,1-y)$. However, without loss of generality, the output of $f_c$ comparison function is assumed to be the first component $z_0$ which represent the probability that $x_0$ is blurrier than $x_1$.

After learning the $f_c$ function, the {\it blurrest} binary relation $R_S$ can be obtained for a burst $S$ of any size. Firstly, $f_c$ is computed for every  pair $\tuple{x} \in S^2$, being obtained as output the probability $\text{P}(\tuple{x})$ that $x_i$ is blurrier than $x_j$. Because a total order over $S$ is required, the $R_S$ definition must meet the \textit{trichotomy law}, which means that for a generic tuple $(x_i,x_j)$ and its reverse $(x_j,x_i)$, also written as $\tuple{x}_{i,j}$ and $\tuple{x}_{j,i}$ respectively, only one of the following holds: $\tuple{x}_{i,j} \in R_S$, $\tuple{x}_{j,i} \in R_S$ or $x_i=x_j$. This implies that $\text{P}(\tuple{x}_{i,j})=1-\text{P}(\tuple{x}_{j,i})$ as considered in $f_c$ training. Then, the probability that $x_i$ is blurrier than $x_j$ is calculated as:

\begin{equation}
\text{P}(\tuple{x}_{i,j})=\frac{f_c(\tuple{x}_{i,j})}{f_c(\tuple{x}_{i,j})+f_c(\tuple{x}_{j,i})}
\end{equation}
and $R_S=\{\tuple{x} \left|\right. \text{P}(\tuple{x}) \geq 0.5 , \forall \tuple{x} \in S^2 \}$. 

In practice, for $R_S$ creation only a subset $\Upsilon_S \subset S^2$ needs to be evaluated on $f_c$. 
%The sufficient subset $\Upsilon_S$ to obtain $R_S$ is determined by the sorting algorithm used for the ranking.
Given a pair $\tuple{x}_{i,j} \in \Upsilon_S$, $R_S$ is created such that if $\text{P}(\tuple{x}_{i,j}) \geq 0.5$ then $\tuple{x}_{i,j} \in R_S$ else $\tuple{x}_{j,i} \in R_S$. This simplification reduces the computation in the worst of the cases to O($n^2$) when used the bubble sort algorithm. The parallel proficiency of current GPUs was used to evaluate $f_c$ for all pairs of $\Upsilon_S$ at once, reducing significantly the execution time for such operations.

Once $R_S$ is derived for a given burst $S$, the totally ordered set $\sburst{S}$ can be obtained. For such, the simplest approach is to compute a score $r_c(i)$ for a given image $x_i$ in the burst as:

\begin{align}
r_c(i) &= \sum_{\tuple{x}_{i,j} \in S^2} \indi{R_S}( \tuple{x}_{i,j})\\
&=\sum_{\tuple{x}_{i,j} \in \Upsilon_S} \indi{R_S}( \tuple{x}_{i,j}) + \sum_{\tuple{x}_{k,i} \in \Upsilon_S} 1 - \indi{R_S}( \tuple{x}_{k,i})
\label{eq:crank}
\end{align}
where $\tuple{x}_{i,j}$ is a simplified notation for the pair $(x_i,x_j)$ and $\indi{R_S}(\tuple{x})$ is the characteristic function over $R_S$. The lowest the value of $r_c(i)$, the sharpest is the image $x_i$ with respect to others images in the burst.

A soft version $r_f$ to compute the score and therefore the sorted set $\sburst{S}$ is proposed as follows:

\begin{align}
r_f(i) &= \sum_{\tuple{x}_{i,j} \in S^2} \text{P}( \tuple{x}_{i,j})\\
&=\sum_{\tuple{x}_{i,j} \in \Upsilon_S} \text{P}( \tuple{x}_{i,j}) + \sum_{\tuple{x}_{k,i} \in \Upsilon_S} 1-\text{P}( \tuple{x}_{k,i})
\label{eq:frank}
\end{align}

Using the statements that $\text{P}(\tuple{x}_{i,j})=1-\text{P}(\tuple{x}_{j,i})$ and  pairs on $\Upsilon_S$ are sufficient to obtain $R_S$, the crisp and soft score computation can be performed as in Eq. \ref{eq:crank} and Eq. \ref{eq:frank} respectively, where only pairs $\tuple{x} \in \Upsilon_S$ are used. The ordinal rank $\gamma$ of $S$ is then obtained such that the rank of $x_i$ is greater than the rank of $x_j$ if $r_f(i)>r_f(j) \left[r_c(i)>r_c(j)\right]$. $\Phi_S$ is the set of elements of $S$ in the order established by $\gamma$.

\subsection{Incremental aggregation}

After obtaining a sorted version $\sburst{S}$ of the burst in an increasing blur order, a deblurring process is carried out. Implementing an information aggregation method, where less blurred information of each image is used to perform the reconstruction, is a successful approach for multi-images deblurring. The classical and simplest form of aggregation is the Fourier Burst Accumulation (FBA) proposed by Delbracio and Sapiro \cite{delbracio2015removing}. This algorithm can be viewed as a lucky image algorithm working in the frequency domain. However, while in lucky methods the burst should contain at least one sharp image or region to obtain a proper reconstruction, the FBA fuses images within a burst by considering their frequency content without the need of sharp frames. The bases of the algorithm are that the motion blur process strongly attenuates some Fourier coefficients while leaving others almost unaltered. If a Fourier coefficient power $p$ is larger in $x$ than its corresponding coefficients power in other frames, then  $p$ has been less attenuated by motion blur. The method is effective even if each image is blurred. A better reconstruction is obtained if the burst is sufficiently time-variant to ensure that each Fourier coefficient has been left unaltered at least once in some neighboring frames \cite{anger2017implementation}. 

\begin{figure}[!tb]
    \begin{center}
        \centering
        
        \includegraphics[width=\linewidth]{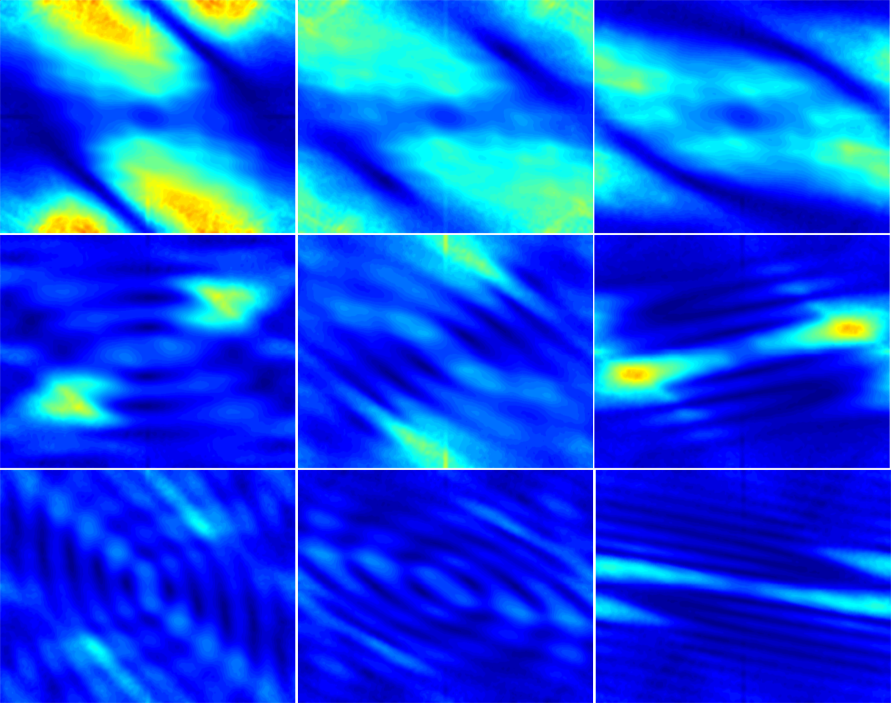}\\
        
    \end{center}
    \caption{FBA weights for nine frames where the top-left image is the sharpest one and the bottom-right is blurrier. The image blur increases from left to right and from top to bottom. The hottest colors represent higher weights while the coolest colors represent the lowest weight values.}
    \label{fig:fabweights}
    %\vspace{-1cm}
\end{figure}

Because in the original FBA proposal the burst order does not matter, the method is implemented using all images at once. Quite the contrary, here an Incremental Fourier Burst Accumulation (IFBA) using only two images at each iteration along with an automatic image selection is proposed. Our method is based on the fact that FBA aggregation usually assigns the higher weights to coefficients of the sharpest images, almost disregarding frequencies in frames with too much blur, see Fig. \ref{fig:fabweights}. Then, from all possible subsets in the powerset of $S$, $\bar{\Phi}_S \in \mathcal{P}(S)$, that can lead to a reconstruction, we are particularly interested in $\bar{\Phi}_S \subset \sburst{S}$ such that if $x_t \in \bar{\Phi}_S$ then $x_{t-1} \in \bar{\Phi}_S$, $x_t$ being the frame $t$ of the sorted burst $\sburst{S}$. In so doing, only the sharpest images of the burst will be used to perform the reconstruction. Our goal then is to automatically find the optimal burst $\bar{\Phi}_{S}^{*} \subset \sburst{S}$ such that the reconstruction $\hat{x} \approx x$, where $x$ is the latent sharp image. In the worst case, that is, all frames are needed to get a significant better reconstruction, the set $\bar{\Phi}_{S}^{*} = \Phi_{S}$ and our method performs similarly to the FBA.

As stated in the problem definition section, an incremental aggregation parametrized by $\omega$ and $p$ is defined as $\hat{x}_t=g_{\omega,p}(x_t,\hat{x}_{t-1})$. The definition of $g_{\omega,p}$ can be obtained directly from the FBA aggregation formulation. For notation simplification let us denote the Fourier Transform of the frame $x_t$ as $v_t=\mathcal{F}(x_t)$ and let $|\bar{v}_t|$ be the mean power of the $c$ image channels. The FBA aggregation function $u_p$ is computed as:

\begin{equation}
u_p(x_0,...,x_l)= \mathcal{F}^{-1} \left[ \sum_{t=0}^{l} \frac{|\bar{v}_t|^p \cdot v_t}{ \sum_{i=0}^{l} |\bar{v}_i|^p} \right]
\label{eq:fba}
\end{equation}

Let's denote $\omega_1=\sum_{i=1}^{l} |\bar{v}_i|^p$. The Eq. \ref{eq:fba} is then expressed as:

\begin{equation}
u_p(x_0,...,x_l)= \mathcal{F}^{-1} \left[ \frac{ |\bar{v}_0|^p \cdot v_0 + \sum_{t=1}^{l} |\bar{v}_t|^p \cdot v_t }{\omega_1 + |\bar{v}_0|^p} \right] 
\end{equation}

giving a recursive expression,

\begin{equation}
u_p(x_0,...,x_l)= \mathcal{F}^{-1} \left[ \frac{ \omega_1 \cdot \mathcal{F}\left(u_p(x_1,...,x_l)\right) +|\bar{v}_0|^p \cdot v_0  }{\omega_1 + |\bar{v}_0|^p} \right] 
\label{eq:rfba}
\end{equation}

Following Eq. \ref{eq:rfba}, the Incremental Fourier Burst Accumulation is obtained as:

\begin{equation}
g_{\omega,p}(x_t,\hat{x}_{t-1})= \mathcal{F}^{-1}\left[ \frac{\omega \cdot \mathcal{F}(\hat{x}_{t-1}) + |\bar{v}_t|^p \cdot v_t }{\omega + |\bar{v}_t|^p } \right]
\label{eq:ifba}
\end{equation}

where $\omega=\sum_{i=0}^{t-1} |\bar{v}_i|^p $ is the Fourier power accumulation of images used to obtain the reconstruction $\hat{x}_{t-1}$.

This incremental formulation of the FBA allow us to efficiently compute an iterative reconstruction with a degradation recognition mechanism. The deterioration measure is needed because the random nature of the blur does not allow a fixed amount of images to be set for obtaining the closest sharp image reconstruction. Considering that $x_t$ is the frame $t$ of the sorted burst $\sburst{S}$, then $\hat{x}_t$ at iteration $t$ will be computed with the $t$ sharpest images of the burst. Because the aggregation of a seriously blurred image can harm the deblurring process, then a reconstruction degradation is measured for automatically stop the IFBA. The relative comparison function $f_c$ is used here to measure a reconstruction degradation as described in previous section. The optimal subset is then build as $\bar{\Phi}_{S}^{*}=\{ x_t \left|\right. \text{P}( (\hat{x}_{t-1},\hat{x}_t) \in R_G ) \geq 0.5 \}$, where $x_t$, $\hat{x}_{t-1}$ and $\hat{x}_t$ are related through Eq. \ref{eq:ifba} as $\hat{x}_t=g_{\omega,p}(x_t,\hat{x}_{t-1})$. 

\subsection{Algorithm implementation}

The Algorithm \ref{alg:ifba} summarize the implementation of the proposed method including the burst sorting (lines 3-8) and the incremental aggregation (lines 9-19) discussed in previous sections. The steps 3-6 of the sorting process were implemented as a single mini batch evaluation over all pairs in $\Upsilon_S$ using GPU parallel computation. The crisp version of the algorithm for implementing Eq. \ref{eq:crank} has a small variation in step 4 for obtaining the indicator function $\indi{R_S}( \tuple{x})$ value, $P_{i,j}= 1$ if $\text{P}(\tuple{x}_{i,j}) \geq 0.5$ else $P_{i,j}= 0$. Note that if $\Upsilon_S=S^2$ the algorithm corresponds to Bubble sort, whereas any other sort algorithm like quick sort can be used instead.

As can be seen in the Algorithm \ref{alg:ifba}, the incremental aggregation steps perform the deblurring using less blurred frames first according to the order in $\sburst{S}$. Because $\omega = 0$ at iteration $t=0$, then $\hat{v}_0=v_0$ (step 13) and $\hat{x}_0=x_0$ (step 15). For values of $t>0$ the aggregation is performed as in Eq. \ref{eq:ifba}. The optimal burst $\bar{\Phi}_{S}^{*}$ is composed by the firsts $t$ images of $\sburst{S}$ until a reconstruction degradation is found (line 16). A Gaussian smoothing $G_{\sigma}$ over the mean magnitude is performed (line 12) selecting $\sigma$ as in \cite{delbracio2015removing}.

\begin{algorithm}[ht!]
    \caption{Incremental burst deblurring}\label{alg:ifba}
    \begin{algorithmic}[1]
        \Procedure{IFBA}{$S$,$p$}
        \State $\textbf{Initialize: } r_f(i) \gets 0, \omega \gets 0,\bar{\Phi}_{S}^{*} \gets \emptyset$
        \Statex \textit{BURST SORTING}
        \For{$\tuple{x}_{i,j} \in \Upsilon_S$}
        \State $P_{i,j} = \frac{f_c(\tuple{x}_{i,j})}{f_c(\tuple{x}_{i,j})+f_c(\tuple{x}_{j,i})}$
        \State $r_f(i) = r_f(i) + P_{i,j}$
        \State $r_f(j) = r_f(j) + (1-P_{i,j})$
        \EndFor
        \State $\gamma=\arg \text{sort}(r_f)$ //Ordinal rank
        \State $\sburst{S}=\{x_{\gamma_0},...,x_{\gamma_l} \}$
        \Statex \textit{INCREMENTAL AGGREGATION}
        \For{$x_t \in \sburst{S}$}
        \State $v_t=\mathcal{F}(x_t)$
        \State $\bar{v}_t=\frac{1}{c}\sum_{k=1}^{c} |v_t^k|$
        \State $\bar{v}_t=G_{\sigma}\bar{v}_t$
        \State $\hat{v}_{t}=\frac{\displaystyle \omega \cdot \hat{v}_{t-1} + |\bar{v}_t|^p \cdot v_t}{\displaystyle \omega + |\bar{v}_t|^p}$
        \State $\omega = \omega + |\bar{v}_t|^p$
        \State $\hat{x}_t=\mathcal{F}^{-1}(\hat{v}_{t})$
        \If {$t > 0 \textbf{ and } \frac{f_c(\hat{x}_{t},\hat{x}_{t-1})}{f_c(\hat{x}_{t},\hat{x}_{t-1})+f_c(\hat{x}_{t-1},\hat{x}_{t})} \geq 0.5$}
        \State $\textbf{break}$
        \EndIf
        \State $\bar{\Phi}_{S}^{*} = \bar{\Phi}_{S}^{*} \cup x_t$
        \EndFor
        \State $\textbf{return } \hat{x}_{t-1}, \bar{\Phi}_{S}^{*}$
        \EndProcedure
    \end{algorithmic}
\end{algorithm}

		\section{Materials and Methods}
\label{sec:results}

To evaluate and validate our approach we conduct several experiments including a comprehensive comparison with state-of-the-art techniques on a real-world dataset, and performance evaluation on a synthetic dataset to test the robustness of our approach with varying image quality of the input sequence.

We trained the sorting network directly on a sequence of unaligned frames featuring large camera shakes. The training is done using only synthetically blurred images. To fulfill the trichotomy law necessary for the sorting, we pass in each train minibatch the tuple $\tuple{x}_{i,j}$ and its reverse $\tuple{x}_{j,i}$. We use RMSProp \cite{hinton212neural} for minimizing the BCE loss function $L=-\left[y\log(z_0)+(1-y)\log(z_1)\right]$. We leave the optimizer's default parameters ($\alpha=0.99$, $\epsilon=10^{-8}$) and the initial learning rate was set to $10^{-5}$. The number of epochs and batch size were $1000$ and $60$ respectively. For kernels generation it was used \textit{anxiety} parameter to $0.008$ \cite{boracchi2012modeling} and trajectory length $m$ spanning in the range $[3:19]$. We further augmented this data with random crops of size $200 \times 200$ and mirroring in every training iteration. Network initialization was made with normally distributed weights using Xavier's method \cite{glorot2010understanding}.

During the inference we pass a pair of frames $\tuple{x}$ and compute the relative ranking as in Algorithm \ref{alg:ifba} using $\Upsilon_S=S^2$ for the step 3. It was assumed a uniform blur in the images, therefore a tile of size $200 \times 200$ was selected from the input frames for faster evaluation. Only images from the validation set of the MS COCO dataset were used in this phase for the synthetic burst generation.

\subsection{Burst sorting evaluation}

For the evaluation of the obtained burst sorting, by means of the comparison function $f_c$, a sequence disagreement distance was used. Given the images ordinal rank $\gamma$ of the sorted burst $\Phi_S$ and the ground truth global score $\delta$, the disagreement is measured with the weighted Kendall $M(\delta,\gamma)$ distance, also denoted as $\tau$ \cite{liu2013no}.

\begin{equation}
M(\delta,\gamma)=\sum_{(i,j) \in D_{(\delta,\gamma)}} |(\max(\delta_i,\delta_j)-\delta_{min})(\delta_i-\delta_j)|
\label{eq:wkendall}
\end{equation}
where $D_{(\delta,\gamma)}$ is the set of pairs $(x_i,x_j)$ whose orders by $\delta$ and $\gamma$ do not agree, and $\delta_{min}$ is the worst Bradley-Terry (B-T) score \cite{bradley1952rank}. For comparison with other metrics whose scores have different scales, we use a normalized version of Eq. \ref{eq:wkendall}, which is defined as $\bar{M}(\delta,\gamma)=M(\delta,\gamma)/M(\delta,-\delta)$, where $M(\delta,-\delta)$ is the maximum mismatch generated by comparing against the reversed ground truth ranking. The distance obtained for the opposite rank $-\delta$ is $\tau_{rev}=1.0$ and $\tau_{rand}=0.5$ for a random guessing function.

For evaluation purposes, we manually sort a set of $30$ synthetically created bursts and compute $\delta$ score from the user labels following the Bradley-Terry model \cite{bradley1952rank}, which is widely used for fitting pairwise comparison results to a global ranking. A comprehensive discussion about the B-T model and how to compute a global score given the pairwise comparison labels can be seen in \cite{liu2013no}\cite{hunter2004mm}. In this work we computed $\delta$ using a quasi-Newton accelerated Minorization-Maximization (MM) algorithm\footnote[1]{http://personal.psu.edu/drh20/code/btmatlab/} for the B-T model \cite{hunter2004mm}. Table \ref{tab:synthetic_sorting} shows the ranking disagreement where lower values are better adequacy to the ground truth rank. Here, $\tau_{f_c}$ corresponds to the weighted Kendall distance of the proposed learned function. The FBA Overall Weights Energy ($OWE$) can also be employed as a comparison criterion because it represents the overall image importance for the FBA deblurring process. A set of non-reference metrics: Cumulative Probability of Blur Detection ($CPBD$) \cite{narvekar2011no}, Gradients of Small magnitudes ($SGrd$) \cite{liu2013no} and Normalize Sparsity ($NSps$) \cite{krishnan2011blind} were also used to rank the burst.

\section{Results}
As can be seen in Table \ref{tab:synthetic_sorting} our ranking function outperforms with high margin the current IQA metrics for recognizing the relative rank of blurred images. The mean execution time for sorting a burst with ten images is approximately $1.30$ seconds for our VGG16-based function. The OWE was second placed but has the limitation that it needs to compute the Fourier Transform for every frame, which is a time-consuming task for high resolutions images. 

\begin{table}[htb!]
\caption{Weighted Kendall $\tau$ distance for $30$ bursts measuring the agreement to the ground truth $\delta$ of our ranking VGG16-based $f_c$ function, $OWE$, $CPBD$, $SGrd$ and $NSps$ metrics. Lower values corresponds to better ground truth agreement.}
    \label{tab:synthetic_sorting}
    \begin{tabular}{cccccccc}
        Burst No.     & $\tau_{f_c}$ (Ours) &    $\tau_{owe}$\cite{delbracio2015removing} &$\tau_{cpbd}$\cite{narvekar2011no}&    $\tau_{sgrd}$\cite{liu2013no}&    $\tau_{nsps}$\cite{krishnan2011blind}\\\hline
        1             & 0.0003&   0.2170&0.0876&    0.3261&    0.0064\\
        2             & 0.0074&    0.0329&0.1164&    0.6196&    0.0537\\
        3             & 0.0120&    0.2173&0.5016&    0.3997&    0.0356\\
        4             & 0.0000&    0.0161&0.0476&    0.6323&    0.0328\\
        5             & 0.0083&    0.1489&0.2257&    0.2930&    0.0588\\
        6             & 0.0051&    0.0432&0.7396&    0.1713&    0.1215\\
        7             & 0.0066&    0.0911&0.1454&    0.2961&    0.4918\\
        8             & 0.0058&    0.2157&0.0643&    0.3250&    0.0052\\
        9             & 0.0222&    0.1529&0.1773&    0.0499&    0.0230\\
        10             & 0.0156&    0.0566&0.2404&    0.0788&    0.3213\\
        11             & 0.0075&    0.2773&0.2293&    0.5058&    0.0371\\
        12             & 0.0027&    0.0673&0.2408&    0.1960&    0.1016\\
        13             & 0.0244&    0.1922&0.6090&    0.0514&    0.0927\\
        14             & 0.0221&    0.0123&0.3927&    0.0140&    0.1193\\
        15             & 0.0093&    0.2372&0.0808&    0.0413&    0.4688\\
        16             & 0.0007&    0.1023&0.2020&    0.0243&    0.3930\\
        17             & 0.0002&    0.0000&0.0171&    0.0638&    0.3587\\
        18             & 0.0015&    0.0908&0.0537&    0.4191&    0.0059\\
        19             & 0.0117&    0.1014&0.5045&    0.0117&    0.0870\\
        20             & 0.0663&    0.2845&0.2230&    0.2786&    0.3533\\
        21             & 0.0046&    0.0547&0.0471&    0.7965&    0.3560\\
        22             & 0.0000&    0.0731&0.2417&    0.1545&    0.0127\\
        23             & 0.0089&    0.0618&0.3373&    0.0117&    0.1891\\
        24             & 0.0190&    0.0831&0.0752&    0.8395&    0.3024\\
        25             & 0.0049&    0.2141&0.2666&    0.0331&    0.0316\\
        26             & 0.0000&    0.0134&0.0863&    0.0132&    0.4964\\
        27             & 0.0017&    0.0234&0.1128&    0.0187&    0.1051\\
        28             & 0.0099&    0.2013&0.3246&    0.4804&    0.0676\\
        29             & 0.0052&    0.0751&0.3035&    0.1438&    0.3036\\
        30             & 0.0323&    0.2245&0.3827&    0.0722&    0.3467\\\hline
        Mean        & \textbf{0.0105}&    0.1194&0.2359&    0.2454&    0.1793\\
    \end{tabular}
\end{table}

\begin{figure}[tbh!]
    \centering
    \includegraphics[width=.9\linewidth]{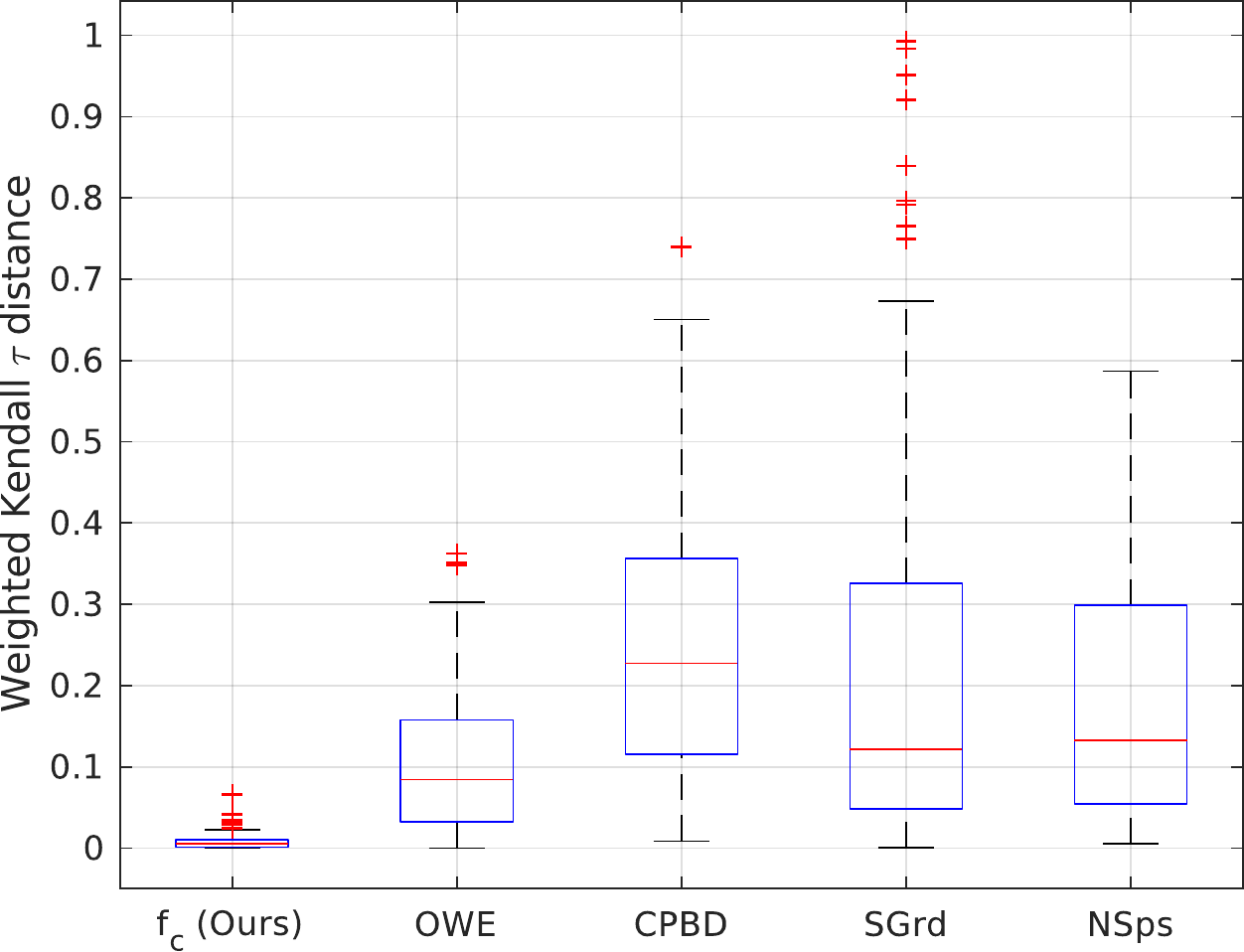}
    \caption{Boxplot of the synthetic blurred burst ranking with the proposed method and $OWE$, $CPBD$, $SGrd$ and $NSps$ metrics.}
    \label{fig:boxplot1}
    %\vspace{-1cm}
\end{figure}

For further analysis, we generate a set of $100$ bursts, and the ground truth score $\delta$ was defined using Eq. \ref{eq:kernel_complex}. The statistical difference between the proposed ranking function and state-of-the-art metrics was probed with the Friedman test \cite{friedman1940comparison} using as significance $\alpha=0.05$. The statistical test resulted in a p-value of $10^{-42}$, strongly rejecting the null hypotheses that all metrics have a similar mean performance. The Nemenyi post-hoc test \cite{nemenyi1962distribution} showed that $f_c$ is significantly different from the other ranking methods, and therefore better. In Fig. \ref{fig:boxplot1} it can be seen the resulting boxplots for analyzed metrics reinforcing the obtained statistical results. Note that the small standard deviation suggests low image content dependence, unlike the $SGrd$ feature.

\begin{figure}[hbt!]
    \begin{center}
        \centering
        \includegraphics[width=\linewidth]{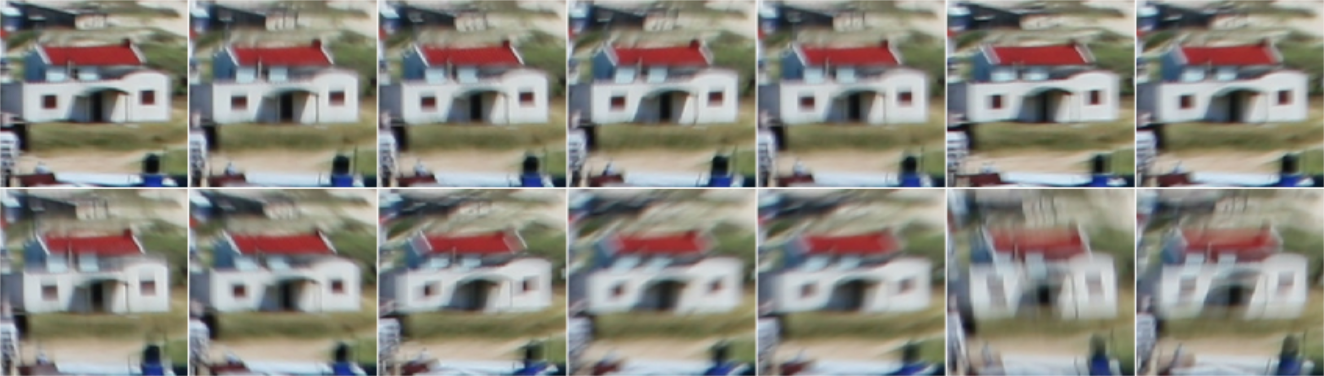}
    \end{center}
    \caption{Obtained burst $\Phi_{S}$ sorted with the proposed $f_c$ comparison function over the real images dataset Pueblo Cabo Polonio \cite{delbracio2015removing}. Images blur increase from left to right and from top to bottom. The obtained weighted Kendall distance was $\tau_{f_c}=0.0040$ when compared with the manually sorted burst.}
    \label{fig:sortingreal}
    %\vspace{-1cm}
\end{figure}

Table \ref{tab:sortime} shows the execution time for a full burst sorting in variable length real datasets. To further improve this computational performance, either a network model compression or Quicksort as a sorting algorithm could be used. Fig. \ref{fig:sortingreal} shows the obtained increasing blur order over the Pueblo Cabo Polonio dataset. The sorting was made in $1.33$ seconds as reported in Table \ref{tab:sortime} and it was not used a prior knowledge about the burst order in network training. The set was sorted by hand and the global score $\delta$ was computed as previously by using the B-T model. The resulting weighted Kendall distance was $\tau_{f_c}=0.0040$ only mistaking the rank of frames 5-6 (first row of the figure), and frames 13-14 (last two images of the second row).

\begin{table}[htb!]
\caption{Sorting time of different length bursts with the proposed VGG16-based function in real images datasets\cite{delbracio2015removing}.}
    \label{tab:sortime}
    \begin{tabular}{lcc}
        Dataset                &Burst length    &Sorting Time (sec)    \\ \hline
        Anthropologie        & 8                & 1.27                \\
        Auvers                & 12            & 1.31                \\
        Bookshelf            & 10            & 1.30                \\
        Lucky Imaging        & 12            & 1.31                \\
        Parking Night        & 10            & 1.30                \\
        Pueblo Cabo Polonio    & 14            & 1.33                \\
        Tequila                & 8                & 1.23                \\
        Woods                & 13            & 1.32                \\ \hline
    \end{tabular}
    \vspace{1mm}
\end{table}

\subsection{Deblurring bursts with fixed frames number}

\begin{figure*}[ht!]
    \begin{center}
        \centering
        \setlength{\tabcolsep}{1pt}
        \small
        \begin{tabular}{p{1.5cm}p{2cm}p{3.6cm}p{3.6cm}p{3.6cm}p{3cm}}
            &&2 Frames&3 Frames&4 Frames&5 Frames\\
            \multicolumn{1}{r}{FBA}&
            \multicolumn{5}{c}{\includegraphics[width=.8\linewidth]{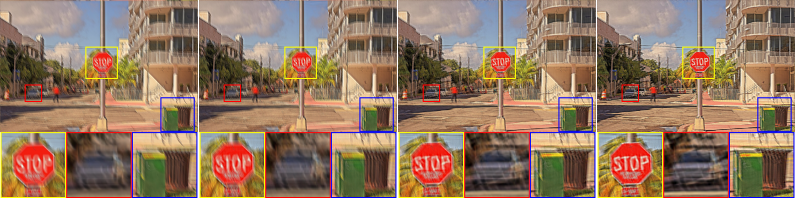}}\\
            
            \multicolumn{1}{r}{IFBA (Ours)}&
            \multicolumn{5}{c}{\includegraphics[width=.8\linewidth]{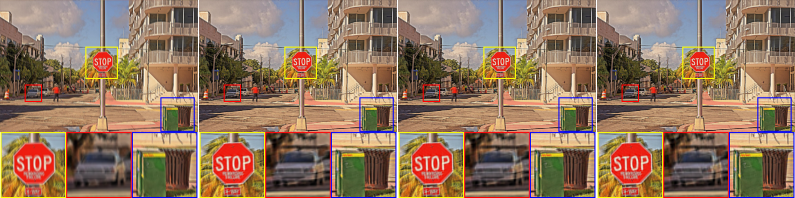}}\\
            
            \multicolumn{1}{r}{FBA}&
            \multicolumn{5}{c}{\includegraphics[width=.8\linewidth]{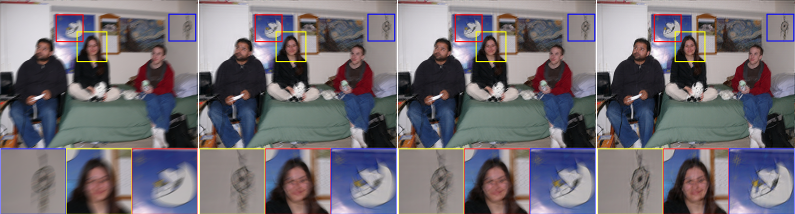}}\\
            
            \multicolumn{1}{r}{IFBA (Ours)}&
            \multicolumn{5}{c}{\includegraphics[width=.8\linewidth]{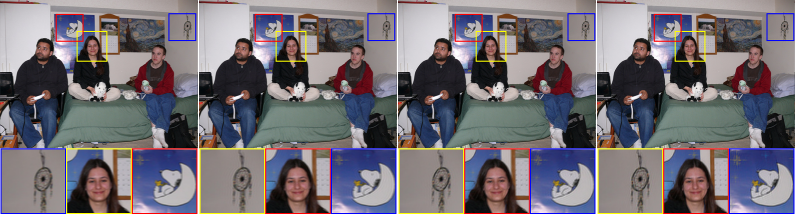}}\\
            
            \multicolumn{1}{r}{FBA}&
            \multicolumn{5}{c}{\includegraphics[width=.8\linewidth]{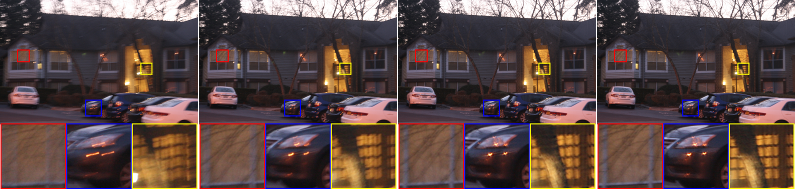}}\\
            
            \multicolumn{1}{r}{IFBA (Ours)}&
            \multicolumn{5}{c}{\includegraphics[width=.8\linewidth]{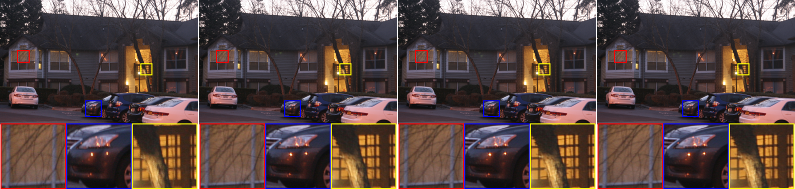}}\\
            
        \end{tabular}
    \end{center}
    \caption{Fixed size aggregation results using FBA and the proposed IFBA. Only four aggregations steps are executed over $S$ for FBA and $\Phi_{S}$ for IFBA. First four rows are synthetic burst and last two rows corresponds to real dataset from \cite{delbracio2015removing}.}
    \label{fig:fixedframes}
    %\vspace{-1cm}
\end{figure*}

In this experiment it is analyzed the burst sorting influence in the reconstruction process. As in \cite{wieschollek2016end}, we test the deblurring method fixing the number of frames used during the aggregation. Fig. \ref{fig:fixedframes} shows the obtained results, visually comparing our approach to the FBA. The first five images of the synthetic dataset were sufficient to obtain a perceptually good reconstruction. A clear improvement is noticed even in the earlier steps of the deblurring. Such a fixed frames number approach is equivalent with multi-image deblurring methods with non-varying temporal size \cite{chakrabarti2016neural,wieschollek2016end,noroozi2017motion} for the aggregation burst length. However, different from others, the reconstruction burst $\Phi_S$ is bounded only in our method, but the input burst $S$ length remains the same. In the last row of Fig. \ref{fig:fixedframes} it is also seen an example of reconstruction on a real image dataset, in which a reconstruction result of quality very close to the final FBA deblurring is obtained within the very first frames from $\Phi_S$, notably improving the FBA on these steps.

\subsection{Reconstruction degradation recognition}

\begin{figure*}[ht!]
    \begin{center}
        \centering
        \setlength{\tabcolsep}{1pt}
        \small
        \begin{tabular}{ccccc}
            Our deblurring of misaligned bursts & Random shot & fastMBD \cite{sroubek2012robust} & FBA \cite{delbracio2015removing} & IFBA (Ours)\\
            \includegraphics[width=0.383\linewidth]{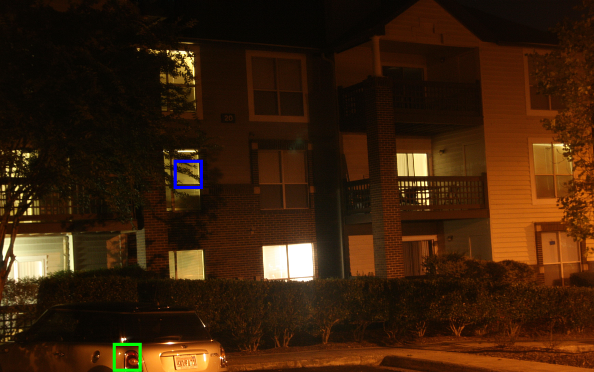}&
            \includegraphics[width=0.12\linewidth]{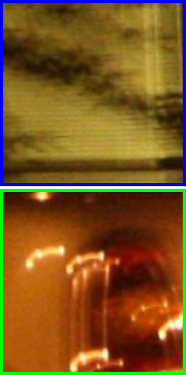}&
            \includegraphics[width=0.12\linewidth]{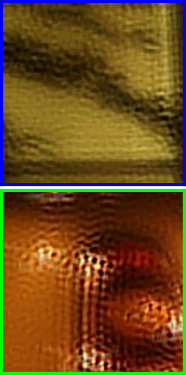}&
            \includegraphics[width=0.12\linewidth]{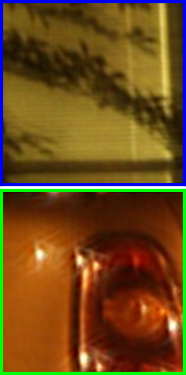}&
            \includegraphics[width=0.12\linewidth]{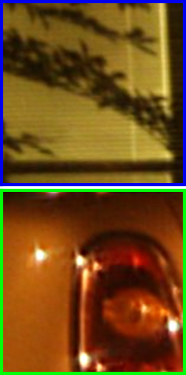}\\

            \includegraphics[width=0.383\linewidth]{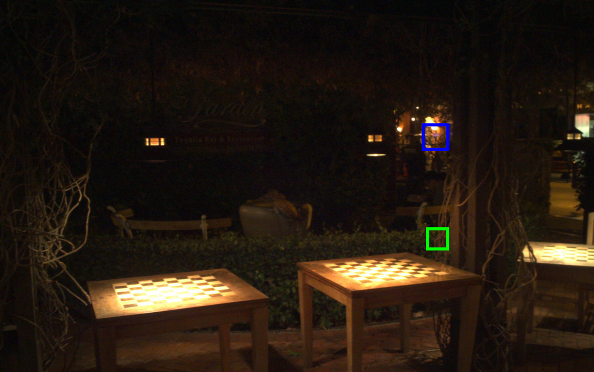}&
            \includegraphics[width=0.12\linewidth]{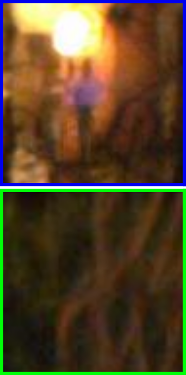}&
            \includegraphics[width=0.12\linewidth]{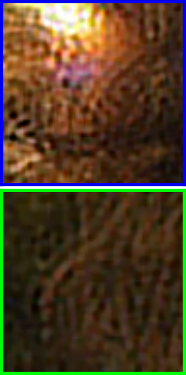}&
            \includegraphics[width=0.12\linewidth]{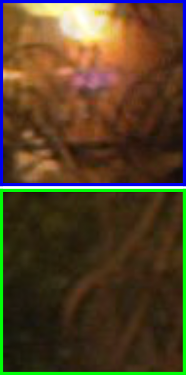}&
            \includegraphics[width=0.12\linewidth]{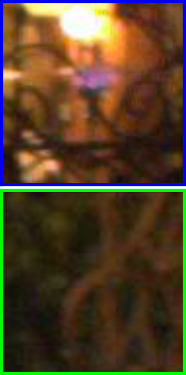}\\

            \includegraphics[width=0.383\linewidth]{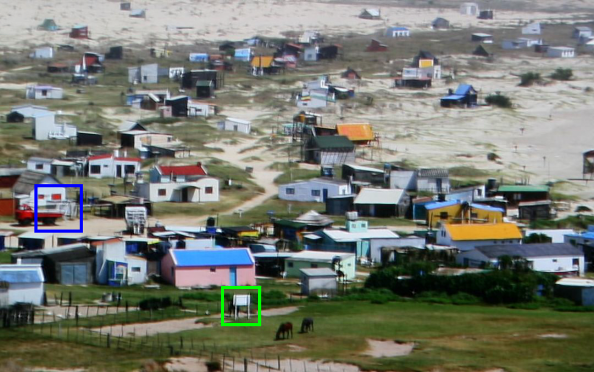}&
            \includegraphics[width=0.12\linewidth]{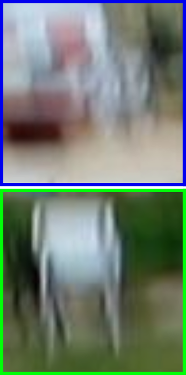}&
            \includegraphics[width=0.12\linewidth]{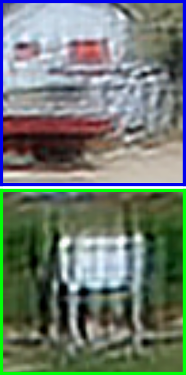}&
            \includegraphics[width=0.12\linewidth]{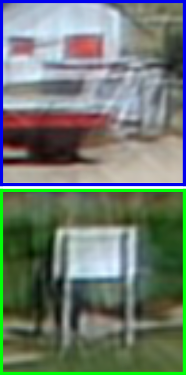}&
            \includegraphics[width=0.12\linewidth]{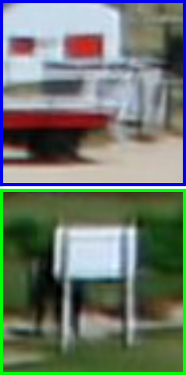}\\
        \end{tabular}
    \end{center}
    \caption{Misaligned bursts deblurring of a shifted version of Ref. \cite{delbracio2015removing} dataset using fastMBD, FBA and our IFBA.}
    \label{fig:deblurring1}
    %\vspace{-1cm}
\end{figure*}

To evaluate the robustness of the automatic image selection mechanism, first, we test it over a misaligned frames dataset. A inaccurate registration is usually obtained when images are too much blurred, as a result features extractors are not able to correctly detect fiducial points, like corners or gradients. The used dataset corresponds to a shifted version of the bursts from \cite{delbracio2015removing}, where half of the frames were randomly selected and shifted. Fig. \ref{fig:deblurring1} depicts some obtained results. A severe degradation in the FBA results is observed because the frames misalignment cause the fusion of non-corresponding frequencies. This mistaken aggregation ends in several artifacts and blurred reconstruction. The proposed incremental FBA, however, fuses all frames within the burst that do not cause a deterioration in the final image as expected. As can be seen in the figure our algorithm surpasses literature methods to a great extent. 

\begin{figure*}[ht!]
    \begin{center}
        \centering
        \setlength{\tabcolsep}{1pt}
        \small
        \begin{tabular}{ccc}
            \multicolumn{3}{c}{Frames sample from the motion blur dataset.}\\
            \includegraphics[width=0.33\linewidth]{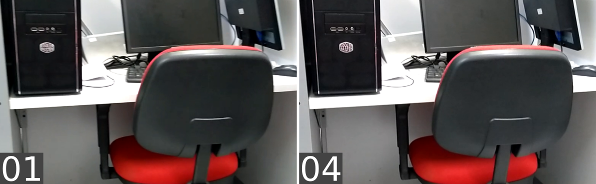}&
            \includegraphics[width=0.33\linewidth]{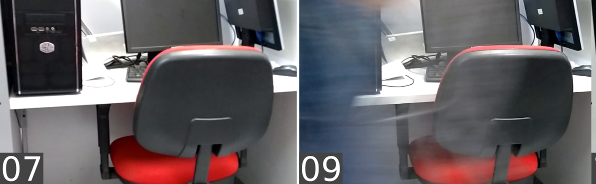}&
            \includegraphics[width=0.33\linewidth]{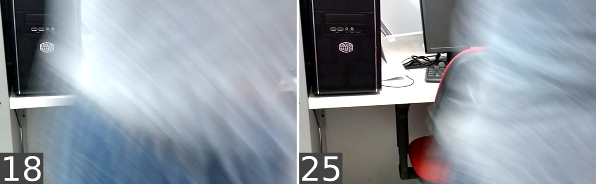}\\

            fastMBD \cite{sroubek2012robust} & FBA \cite{delbracio2015removing} & IFBA (Ours)\\
            \includegraphics[width=0.33\linewidth]{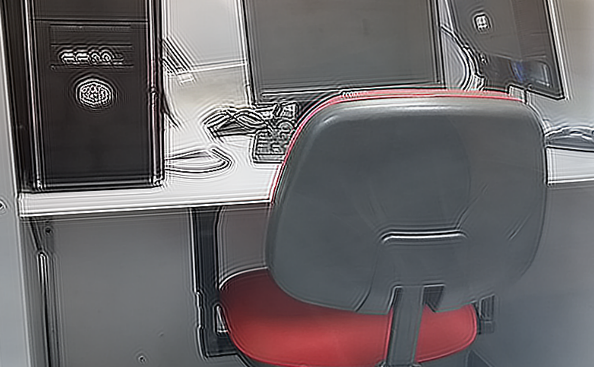}&
            \includegraphics[width=0.33\linewidth]{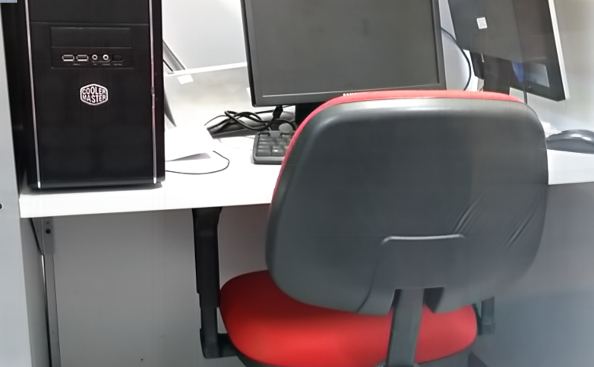}&
            \includegraphics[width=0.33\linewidth]{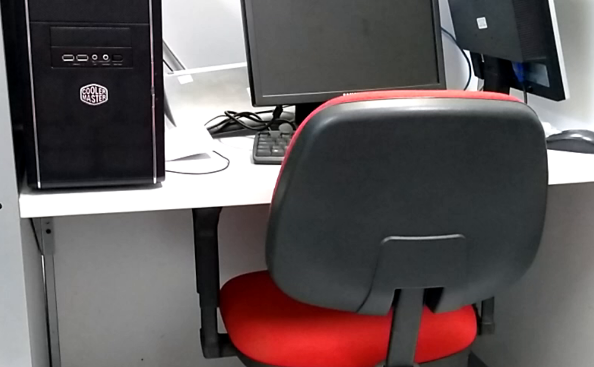}\\

        \end{tabular}
    \end{center}
    \caption{Deblurring results over a motion blur dataset with partial and full occlusion of the scene. First row shows a numbered sample of the frames within the burst. The obtained reconstructions through our method and other approaches can be observed in the second row.}
    \label{fig:deblurring2}
    %\vspace{-1cm}
\end{figure*}

Another experiment was performed over a dataset of real motion-blurred images. With this purpose, we recreate another common real-life situation where a burst is captured, and a moving object appears in the scene partially or entirely occluding the target object. A total of 27 frames were captured under camera shake and motion blur. The registration was performed using SURF \cite{bay2006surf} to extract the features and RANSAC \cite{fischler1981random} for matching, followed by a similarity transformation estimation. Fig. \ref{fig:deblurring2} displays an example of the captured frames and the obtained results. It can be observed brightest reconstruction with fastMBD and FBA because all the frames within the burst in the reconstruction process were selected. Ringing artifacts and blurred regions are also seen whenever all images in the burst are considered. Nevertheless, one more time our proposal performs an effective selection of images resulting in a better fusion by using only the first seven frames of the burst.

		\section{Conclusions}
		\label{sec:conclusion}
		
		In this work a new relative ranking method for frames within a burst using a CNN as a comparison function is proposed. An incremental aggregation with reconstruction degradation recognition to fuse images that do not cause a drop in the reconstruction quality is also introduced. We conducted several experiments for validation of the proposed burst sorting and incremental aggregation. It was demonstrated the superiority of our approach when compared to other similar methods in the wild. The burst sorting algorithm shows a good agreement with the ground truth rank, while outperforming by a large margin other literature metrics. We also improved the results over misaligned and out-of-context frames through the use of our incremental aggregation.
		
        \section*{Acknowledgment}
		\addcontentsline{toc}{section}{Acknowledgment}
        We thank financial support from the Brazilian funding agencies
  FACEPE, CAPES and CNPq. This work was supported by the research cooperation project between Motorola Mobility LLC (a Lenovo Company) and the Center for Informatics of the Federal University of Pernambuco. The authors would also like to thank Leonardo Coutinho de Mendon\c{c}a, Alexandre Cabral Mota and Rudi Minghim for valuable discussions.

		\bibliographystyle{IEEEtran}
		\bibliography{refs}

		
%		\begin{IEEEbiography}{Fidel Alejandro Guerrero Pe\~{n}a}
%			Biography text here.
%		\end{IEEEbiography}
%		
%		% if you will not have a photo at all:
%		\begin{IEEEbiographynophoto}{John Doe}
%			Biography text here.
%		\end{IEEEbiographynophoto}
%		
%		% insert where needed to balance the two columns on the last page with
%		% biographies
%		%\newpage
%		
%		\begin{IEEEbiographynophoto}{Jane Doe}
%			Biography text here.
%		\end{IEEEbiographynophoto}
		
		%\vfill
		

	\end{document}